\documentclass{SciPost}
\usepackage{graphicx} 
\usepackage{amsmath}
\usepackage{amssymb}
\usepackage[table]{xcolor}
\usepackage{booktabs} 

\hypersetup{
    colorlinks,
    linkcolor={red!50!black},
    citecolor={blue!50!black},
    urlcolor={blue!80!black}
}

\usepackage[bitstream-charter]{mathdesign}
\DeclareSymbolFont{usualmathcal}{OMS}{cmsy}{m}{n}
\DeclareSymbolFontAlphabet{\mathcal}{usualmathcal}

\fancypagestyle{SPstyle}{
\fancyhf{}
\lhead{\colorbox{scipostblue}{\bf \color{white} ~SciPost Physics Community Reports }}
\rhead{{\bf \color{scipostdeepblue} ~Submission }}

\fancyfoot[C]{\textbf{\thepage}}
}

\definecolor{TuneColor}{HTML}{2563EB}  
\definecolor{DepthColor}{HTML}{DC2626}  
\definecolor{WidthColor}{HTML}{7C3AED}   

\newcommand{\widthctl}[1]{\textcolor{WidthColor}{#1}}   
\newcommand{\depthctl}[1]{\textcolor{DepthColor}{#1}}   
\newcommand{\tunable}[1]{\textcolor{TuneColor}{#1}}     

\begin{document}

\pagestyle{SPstyle}

\begin{center}{\Large \textbf{\color{scipostdeepblue}{
Statistical Properties of Training \& Generalization\\
}}
\normalsize Part of the VERaiPHY Initiative}\end{center}

\begin{center}\textbf{
Itay Lavie\textsuperscript{1$\star$},
Noam Levi\textsuperscript{2$\star$} and
Yonatan Kahn\textsuperscript{3$\circ$}
}\end{center}

\begin{center}
{\bf 1} John A. Paulson School of Engineering and Applied Sciences, Harvard University, Cambridge, MA, USA
\\
{\bf 2} Tel Aviv University, Tel Aviv, Israel
\\
{\bf 3} Department of Physics, University of Toronto and Vector Institute, Toronto, ON, Canada
\\[\baselineskip]
$\star$ Leading authors\quad
$\circ$ Advisor
\end{center}

\section*{\color{scipostdeepblue}{Abstract}}
\textbf{\boldmath{%
Deep learning has managed to evade numerous intuitions from classical statistics to achieve unprecedented performance on a number of real-world tasks. In this article, we investigate the key features and surprises of deep learning from a physics-informed perspective, taking care to point out and justify where possible the many choices inherent in constructing a deep learning model. In particular, we review the phenomenon of neural scaling laws and discuss their interplay with the constraints and inductive biases which may be present when applying machine learning to problems in physics.
}}

\vspace{\baselineskip}



\vspace{10pt}
\noindent\rule{\textwidth}{1pt}
\setcounter{tocdepth}{2}
\tableofcontents
\noindent\rule{\textwidth}{1pt}
\vspace{10pt}


The principles of probability and statistics, applied to parameter estimation and hypothesis testing, form the bedrock of any data-driven scientific inquiry. However, the application of these principles to deep neural networks (DNNs) reveals a landscape where classical intuition, developed in the context of lower-dimensional and often convex problems, can be misleading. The sheer scale of modern models (whether measured in terms of parameters, dataset size, computing power, or all of the above) introduces qualitatively new phenomena that demand dedicated study and possibly new theories; in the words of Anderson: ``more is different"~\cite{andersonMoreDifferent1972}.

This article aims to explore the statistical properties that govern the training dynamics and generalization capabilities of these powerful models. From a statistical perspective, the training of a neural network is a stochastic process of navigating a complex, high-dimensional, and non-convex loss landscape to find a set of parameters, $\boldsymbol{\theta}$, that minimizes a chosen loss function $\mathcal{L}(\boldsymbol{\theta})$. This process is analogous to parameter estimation, but the model $f(\mathbf{x}; \boldsymbol{\theta})$ is a highly flexible, non-linear function approximator which has an iterative layer-by-layer structure, leading to useful theoretical and empirical properties as we will discuss. The ultimate goal, \emph{generalization}, is the model's ability to make accurate predictions on new data drawn from the same underlying distribution, the crucial step that allows us to infer physical principles from finite datasets.

Our exploration is structured to build from the ideal to the practical.
\begin{itemize}
    \item We begin in Sec.~\ref{sec:universal} by examining the \textbf{universal aspects of modern deep learning} that emerge in the highly over-parameterized regime, where the number of model parameters far exceeds the number of training samples. Here, we discuss surprising phenomena such as benign overfitting, the double descent of the test error, and the empirical discovery of neural scaling laws, which suggest that in an idealized limit, performance scales predictably with model size, data, and compute.
    
    \item In Sec.~\ref{sec:design}, we examine how these universal behaviors are modulated by \textbf{hyperparameter choices}. The specific learning dynamics and the final learned model are critically sensitive to decisions regarding the architecture, initialization scheme, optimizer, and learning rate. Understanding these effects is essential for bridging the gap between theoretical properties and practical performance.
    
    \item Finally, in Sec.~\ref{sec:constraints}, we confront the reality of \textbf{learning under fundamental constraints}. These constrained scenarios are directly relevant to pressing challenges in physics, from searches for rare signals in high-energy physics (data-limited) and the deployment of models on resource-constrained detector hardware (parameter- and compute-limited), to constructing efficient emulators which reproduce the statistics of computationally-expensive cosmological simulations. We will discuss how these constraints break the assumptions of the idealized scaling regime and motivate specific strategies, such as the incorporation of physics-based inductive biases, to achieve robust performance.
\end{itemize}

Through this structured exploration, we aim to build a bridge from foundational theory to the practical, and often surprising, realities of applying deep learning in the physical sciences, providing a guide for the scientifically sound and informed use of these tools. This article contributes to VERaiPHY (Validation \& Evaluation for Robust AI in PHYsics), a PHYSTAT review series establishing verification and validation standards for machine learning across particle physics, astrophysics, and cosmology.

\section{Universal Aspects of Modern Deep Learning}
\label{sec:universal}

While the architectures and applications of deep learning are vast and varied, a set of universal properties has been observed to emerge in the modern regime of large-scale models and datasets. These empirical observations, often established in the idealized limit of infinite parameters, data, and compute, form the foundation of our current understanding. They challenge classical statistical learning theory and provide a baseline from which we can understand the effects of practical constraints, as we will discuss in later sections.

\subsection{Classical vs. Deep Learning}

A principal distinction between classical machine learning and modern deep learning resides in the geometry of their respective optimization problems. Classical methods, such as linear regression, logistic regression, and Support Vector Machines (SVMs), are often formulated as \emph{convex optimization} problems. The loss function $\mathcal{L}(\boldsymbol{\theta})$ over the parameter space $\boldsymbol{\theta}$ possesses a single global minimum, guaranteeing that gradient-based algorithms can provably converge to the optimal solution.

In stark contrast, the loss function of a deep neural network is profoundly \emph{non-convex}. This property is an inevitable consequence of the compositional structure of deep neural networks, which are composed of building blocks of linear or nonlinear operations iterated across multiple layers. The resulting loss landscape is extraordinarily complex, featuring a vast number of critical points, including numerous local minima, saddle points, and flat plateaus~\cite{dauphin_identifying_2014}. Empirical and theoretical analyses of deep-network loss surfaces support this qualitative picture, including the prevalence of saddle points in high dimensions and the relative benignity of many local minima in certain model classes~\cite{dauphin_identifying_2014,choromanska_loss_2015,kawaguchi_deep_2016}. From a classical optimization perspective, this landscape should be prohibitively difficult to navigate. One would expect simple, local optimizers such as Stochastic Gradient Descent (SGD) to become trapped in suboptimal local minima, preventing the model from achieving high performance.

Surprisingly, this does not seem to be the case. The remarkable empirical success of SGD and its variants in this complex landscape has been a central topic of theoretical inquiry \cite{choromanska_loss_2015}. While a complete picture is still emerging, two key insights help explain this phenomenon:
\begin{enumerate}
    \item \textbf{The blessing of dimensionality:} In high-dimensional optimization, as is common in modern deep learning, it is believed that ``bad" local loss minima are rare, and instead saddle points become the obstacle to optimization. Fortunately, stochasticity in SGD, along with techniques such as momentum\cite{polyak_some_1964,nesterov_introductory_2004,sutskever_importance_2013}, provides an effective mechanism for escaping these saddle points~\cite{jin_agd_2018}.
    \item \textbf{Over-parameterization and loss landscape topology:} In the highly over-parameterized regime (where the number of parameters far exceeds the number of training samples), the loss landscape appears to become smoother. It is conjectured that the minima are not isolated points but form large, connected, low-loss subspaces, making it easier for an optimizer to find a good solution \cite{draxler_essentially_2018}.
\end{enumerate}

It is worth noting that the ``pathologies'' of non-convex optimization are often less severe in modern deep learning than worst-case theory might suggest. 
First, a large class of critical points are \emph{saddles} rather than poorly generalizing minima, implying that in realistic scenarios, gradient descent will avoid them. 
Second, in certain overparameterized regimes, one can prove that (stochastic) gradient methods reach global minima (or near-global minima) at polynomial rates with explicit dependencies on learning rate, width, and data separation~\cite{du_globalminima_2019,arora_finegrained_2019,arora_implicitaccel_2018}.
These results connect optimization dynamics directly to representation and generalization, and they help rationalize why momentum and carefully tuned (sometimes large) step sizes can be effective in practice: in benign landscapes, acceleration can improve the transient rate of progress while overparameterization mitigates spurious bad minima~\cite{polyak_some_1964,nesterov_introductory_2004,sutskever_importance_2013}. 
Nevertheless, even when the conditions in these proofs are violated, deep learning models often still successfully generalize, leaving a substantial gap between what can be shown theoretically and what is experimentally shown to hold in practice.

The standard SGD update rule chooses a subset of the data called a mini-batch $\mathcal{B}$ and runs gradient descent on it:
\begin{equation}
    \boldsymbol{\theta}_{t+1} = \boldsymbol{\theta}_t - \eta \nabla_{\boldsymbol{\theta}} \mathcal{L}_{\mathcal{B}}(\boldsymbol{\theta}_t),
    \label{eq:sgd_update}
\end{equation}
where $\eta$ is the learning rate. This cycle repeats until the model has seen the entire dataset exactly once -- a milestone known as an ``epoch", where training often consists of multiple epochs.  Despite its simplicity, this algorithm, when applied to deep models, reliably finds solutions that not only minimize the training loss but also generalize well to unseen data, yet another surprising property which we explore in Sec.~\ref{sec:generalization}. It is worth noting that, while almost all modern machine learning optimizers (many of which we review in Sec.~\ref{sec:optimizers}) are some variant of SGD, physicists have developed novel physics-inspired optimizers by analogy to interesting dynamical systems, for example Refs.~\cite{DeLuca:2022brp,DeLuca:2023yld,DeLuca:2025ruv}. While we will not consider these optimizers further in this work, their behavior with respect to the phenomena we identify is a ripe area for further study.

\subsection{Generalization and Benign Overfitting}
\label{sec:generalization}

The classical paradigm for model generalization is the bias-variance tradeoff, which posits a U-shaped curve for the test error as a function of model complexity. Initially, simple models (low complexity) ``underfit", and their error is dominated by high bias. As complexity increases, bias decreases and test performance improves. However, beyond an optimal point, the ``sweet spot", the model becomes overly complex and begins to ``overfit" the training data, capturing spurious noise. This leads to high variance in the model parameters and a degradation in test performance.

\begin{figure}
    \centering
    \includegraphics[width=0.5\linewidth]{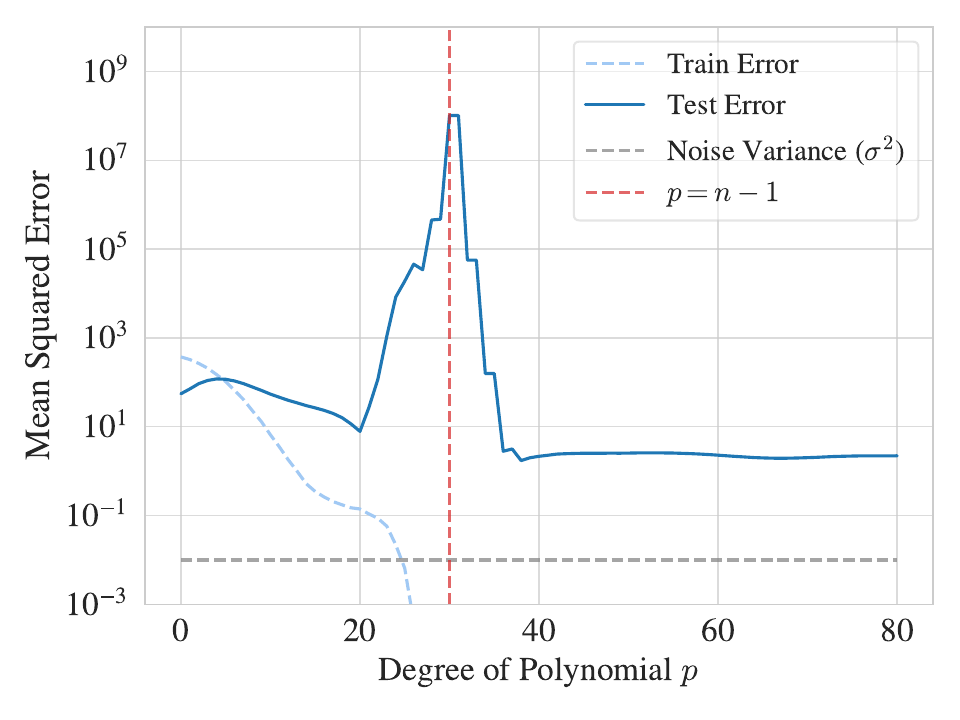}
    \caption{Bias-variance trade-off and double descent. The test error at first decreases as the number of parameters (polynomial degree) is increased, then reaches a maximum when the number of parameters equals the number of samples. After this peak, a second descent begins where more parameters yields lower error.}
    \label{fig:polynomial_double_descent}
\end{figure}

Modern deep learning has revealed a striking departure from this classical picture through the \emph{double descent} phenomenon \cite{opper_statistical_1996,advani_high-dimensional_2017,Geiger18,belkin_reconciling_2019, nakkiran2021deep,hastie_surprises_2020}. Here, the test error follows a more complex trajectory. As model complexity (e.g., number of parameters) increases, the test error initially follows the classical U-shaped curve. This trend continues up to the \emph{interpolation threshold}, where the model has sufficient capacity to perfectly fit the training data. Beyond this point, in the highly over-parameterized regime, the test error, counter-intuitively, begins to decrease again. This leads to a state of \emph{benign overfitting}, where a model can achieve zero training error (perfect interpolation) yet still exhibit excellent generalization to test data. An example in linear regression setting is shown in Fig.~\ref{fig:polynomial_double_descent} (see more details in~\ref{subsec:linear_regression}).

In relatively simple models, aspects of benign overfitting can be characterized rigorously: for example, in overparameterized linear regression the minimum-$\ell_2$-norm interpolator can achieve near-optimal prediction risk under explicit conditions on the data covariance (effective-rank criteria)~\cite{bartlett_benign_2020,tsigler2023benign}, and related analyses extend to certain kernel and random-feature estimators.

This behavior is intimately linked to the concept of \emph{inductive bias}, which refers to the learning assumptions that dictate how a model deduces a function from the data. The inductive bias can be anything from a rigid nearest-neighbor algorithm to an equivariance with respect to some symmetry, or (to give a physically-inspired example) a preference towards functions with low-order spherical harmonics. From a Bayesian perspective, an inductive bias is a choice of prior over the space of functions. In neural networks, the choice of inductive bias is often implicit in design choices like architecture, width, depth, regularization, etc. From the optimization side, such an inductive bias is often thought of as an \emph{implicit regularization}. The choice of optimization algorithm itself imparts a bias on the type of solution found. It is hypothesized that among the infinite possible solutions that can perfectly interpolate the training data, SGD-based methods preferentially converge to "simpler" ones~\cite{soudry2018implicit, gunasekar_implicit_2018}. For linear regression, SGD, when initialized at the origin, is known to converge to the minimum $\ell_2$-norm solution, which is a form of Tikhonov (or ridge) regularization (see e.g. ~\cite{hastie_surprises_2020}).

\subsubsection{Example: Linear Regression}
\label{subsec:linear_regression}
Surprisingly, architectural inductive bias, optimizer implicit bias and double descent phenomena can all be seen in linear regression (i.e. simple polynomial fit)~\cite{belkinUnderstandDeepLearning2018,wilsonDeepLearningNot2025,hastie_surprises_2020}. In this section we work through an explicit example, showing how all these concepts are made manifest, and draw conclusions for large model training.

\begin{figure}
     \centering
    \includegraphics[width=0.9\linewidth]{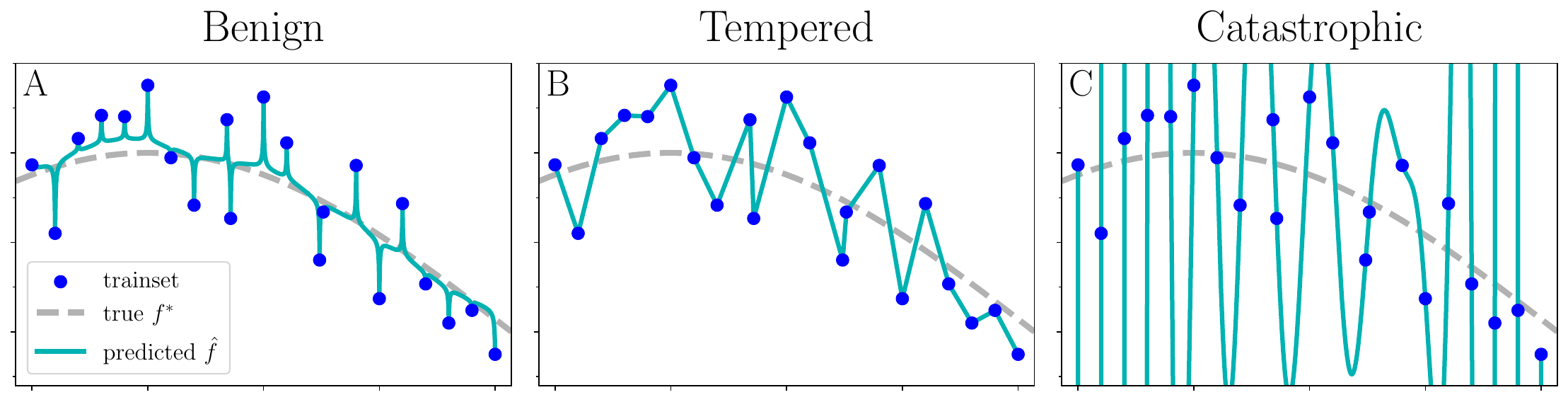}
    \caption{An illustration of benign overfitting, reproduced from~\cite{mallinar2022benign}. Overfitting can be benign ({\bf left}) where the predictor is correct in almost all points, and simultaneously interpolates the training data perfectly. It can also be tempered ({\bf center}) where the predictor gets the function approximately correct everywhere, but overfitting clearly hurts generalization. Finally, overfitting can be catastrophic ({\bf right}), where the data is interpolated perfectly but the prediction is wrong everywhere. The latter case is the situation at the peak of the double descent.}
    \label{fig:benign_overfitting}
\end{figure}

Consider learning a non-linear function like $\tanh$ from $n$ noisy samples, using the first $p$ Legendre polynomials. 
In the under-parametrized (classic) $p < n$  regime, the problem is overdetermined. The unique solution is trivially the min-norm solution, and we do not need to concern ourselves with inductive biases. In this regime, the most natural question is how many parameters to use. Classical learning theory, which describes this overdetermined regime well, presents us with a tradeoff between bias and variance. If we have very few parameters in the model, it won't be able to describe the ground truth function, so increasing the number of parameters decreases the bias. On the other hand, if we have too many parameters, the model becomes sensitive to the noise in the data and overfits it, increasing the variance. The result is a U-shaped curve where on the left the error is dominated by the bias and on the right by the variance. Beyond that variance ascent lies the second descent (hence the name ``double descent") of the over-parametrized regime.

In the over-parameterized (modern) $n < p$  regime, the problem is underdetermined, and one must add additional constraints in order to find a concrete solution. As mentioned above, in this setting, GD has a well-understood implicit bias towards reaching the min-norm solution among all the possible solutions to the least-squares problem
\begin{equation}
    \hat{\beta} = {\arg\!\min_\beta} \sum_{\mu=1}^n \left( X_{\mu,i} \beta_i -y_\mu \right)^2
\end{equation}
with $X\in \mathbb{R}^{n \times p}$ the design matrix, $\beta \in \mathbb{R}^p$ the model parameters and $y = X \beta^* + \sigma^2 \xi$ the target, where $\xi \sim N(0,1)$. The solution thus heavily depends on the norm of the features on the data - features with larger norm will be preferred as they necessitate a smaller $\beta$ and thus preferred by GD; this is the architectural inductive bias. More generally, the eigenvalues of the feature covariance $X^T X$ control the learnability of their corresponding eigenvalues - the higher the eigenvalue the easier the feature is to learn.  In our demonstration, we will use data uniformly sampled on $[-1,1]$, which means the covariance is diagonal in the familiar Legendre polynomials
\begin{equation}
    [X^T X]_{nm} \approx \int x_n x_m \, p(\vec{x}) \, d\vec{x} = \delta_{nm} \frac{2}{2n+1};
\end{equation} 
implying $\lambda_n \sim n^{-1}$. We see that some inductive bias is already baked into the standard definition of the Legendre polynomials: obtaining a unit-norm polynomial of degree $10$ requires a $\beta$ component that is $10$ times larger than a unit-norm constant. This architectural inductive bias, together with the implicit bias of GD, results in a preference for explaining the data using low-degree polynomials. Put simply and somewhat cartoonishly, combining GD with the standard definition of the Legendre polynomials leads to a procedure that effectively fits the lower-degree components first, using higher-degree components only to explain the residuals of that fit. The overfitting that occurs in this regime is said to be ``benign" because the important structure in the data is captured correctly, degree by degree, while the spurious high-degree polynomials are used sparingly and only to interpolate the small amount of noise in the data. This is illustrated in the left panel of Fig~\ref{fig:benign_overfitting}.

Remarkably, these results carry over almost as-is to neural network in the lazy~\cite{cohen_learning_2021,canatar_spectral_2021,simon_eigenlearning_2023,lavie_demystifying_2025,karkadaPredictingKernelRegression2025} NTK/NNGP regime (see Section~\ref{subsec:lazy_rich} below for discussion of rich and lazy learning). These results also play and important role in the rich regime, where the feature covariance is replaced by a kernel that adapts to the target function~\cite{seroussi_separation_2023,rubin_kernels_2025,ringel_applications_2025,lauditi_adaptive_2025}.

\begin{figure}
    \centering
    \includegraphics[width=0.45\linewidth]{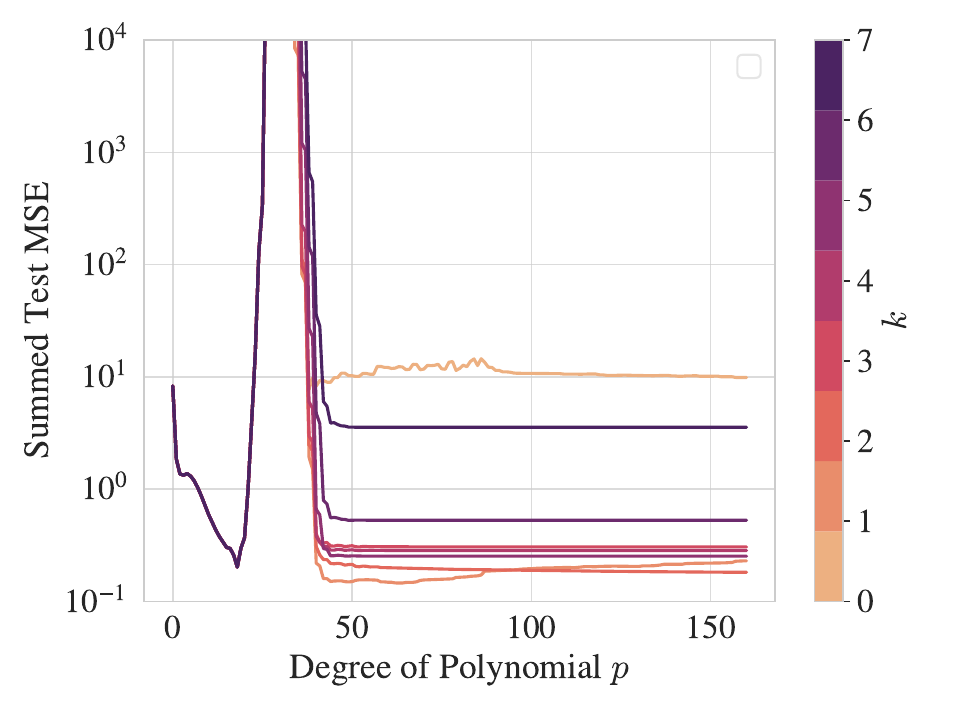}
    \includegraphics[width=0.45\linewidth]{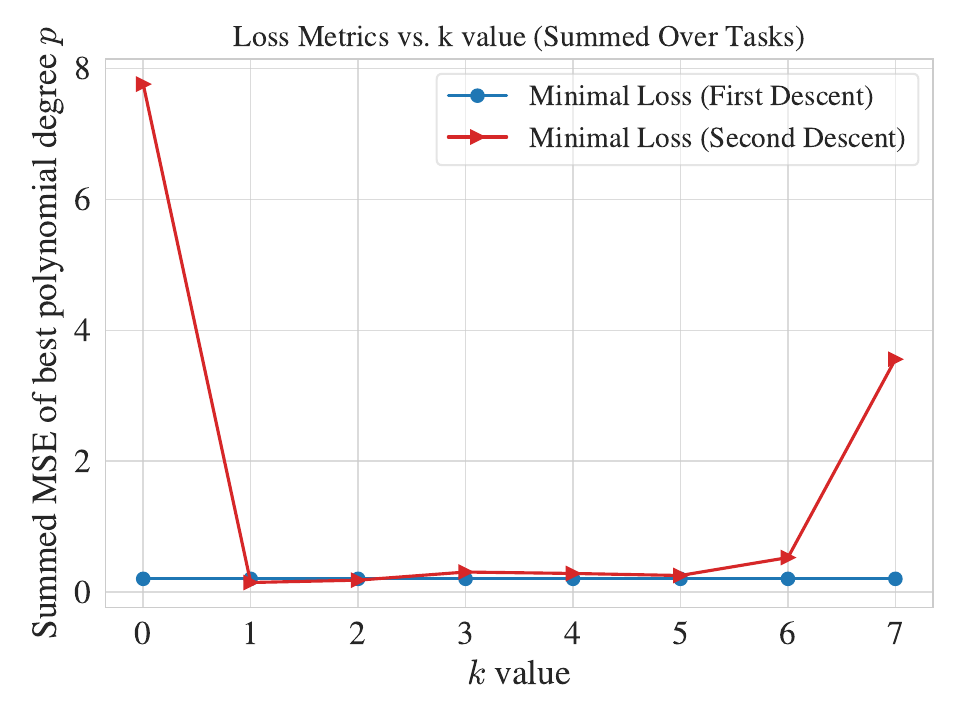}
    \caption{{\bf Left}: Test error summed over three different target functions as a function of the polynomial degree $p$. Colors indicate different inductive bias parameters $k$. The under-parametrized regime is highly sensitive to the parameter count, while the over-parametrized regime is largely insensitive to it. {\bf Right:} Minimal (w.r.t. $p$) test error summed over three different target functions achieved in the under parameterized (blue) and over-parameterized (red) regimes, as a function of the inductive bias parameter $k$. The under-parametrized regime (first descent) is invariant to the inductive bias parameter $k$, while the over-parameterized regime has a very wide minimum in the inductive bias parameter $k$. Code to reproduce these plots may be found \href{https://colab.research.google.com/drive/1TotYZpWa4YudvzUuvWwyn_iQkDdA1ToU?usp=sharing}{here}.}
    \label{fig:polynomial_inductive_bias}
\end{figure}

Fig.~\ref{fig:polynomial_inductive_bias} (left) shows the train and test error when fitting Legendre polynomials as a function of the maximal degree allowed $p$. In the test error we can see first the U-shaped part, followed by the second descent. On the other hand, the train error monotonically decreases with $p$, vanishing at the interpolation threshold.  

We can further explore the effect of the architectural inductive bias by changing the normalization of the Legendre polynomials such that $[X^T X]_{nm} \propto \delta_{nm} n^{-k}$ (where the standard normalization is $k=1$), thereby changing the strength of the preference towards low-degree polynomials that is built-in to our choice of basis. Here we can look for the optimal ``architecture" for a specific task, or consider many different tasks and try to find a universal ``architecture" that will work well for a family of functions simultaneously. Fig.~\ref{fig:polynomial_inductive_bias} shows the test loss averaged for three different ``tasks'': $\arctan(x), \text{ReLU}(x), \sum_{t=1}^{30} \sqrt{t} P_t (x)$, where $P_t$ is the $t$'th Legendre polynomial. We see that in the under-parameterized regime, one has to fine-tune $p$ in order to get good performance, while in the over-parameterized regime, any $p$ large enough would yield good results. The mirror image is shown in the right panel where we consider the loss as a function of $k$. The choice of $k$ has no effect in the under-parameterized regime, but can lead to performance gains or losses in the over-parameterized regime; nevertheless, the performance is far less sensitive to the choice of $k$ than to the choice of $p$.  

When considering costly models like LLMs that are trained only once, minimizing uncertainty and tweaking by choosing the over-parameterized regime is the only feasible approach.

\subsubsection{Lazy Learning and the Kernel Limit of Neural Networks} \label{subsec:lazy_rich}
A useful bridge between the linear-regression example above and fully nonlinear neural networks is obtained by taking an \emph{infinite-width limit}. As the hidden-layer widths are taken to infinity under standard normalization, the prior over network outputs at random initialization converges to a Gaussian process characterized by the \emph{Neural Network Gaussian Process} (NNGP) kernel~\cite{Neal1996,Lee2017,matthews2018gaussian}. Simultaneously, gradient descent training \emph{linearizes}: the network parameters barely move from their initial values, and the dynamics become equivalent to kernel regression with a deterministic, time-invariant \emph{Neural Tangent Kernel} (NTK)~\cite{jacot_ntk_2018}. Because the internal representations are essentially frozen throughout training, this is known as the \emph{lazy} regime~\cite{chizat2019lazy}: generalization reduces precisely to the spectral alignment problem illustrated in Sec.~\ref{subsec:linear_regression}, now with the NTK eigenspectrum playing the role of the feature covariance $X^\top X
$\footnote{Note that the NTK is a kernel, a sample by sample matrix, whereas the design matrix is a covariance, a feature by feature matrix, the two views are equivalent as can be seen by the push-through matrix identity $(I+UV)^{-1} U = U (I+VU)^{-1}$.}.
In practice, however, networks often operate far from this limit. In the complementary \emph{rich} (or \emph{feature learning}) regime, the internal representations adapt during training—the effective kernel evolves toward the target function, which can substantially improve sample efficiency when the initial kernel is poorly aligned with the task. Whether an infinitely-wide network operates in the lazy or rich regime is governed by the relationship between initialization scale, learning rate, and network width; making this relationship precise and systematic is the main goal of the parameterization strategies we discuss in Sec.~\ref{sec:design}.

\subsection{Neural Scaling Laws}
\label{sec:scalinglaws}

Having established that over-parameterization is not necessarily detrimental, but can in fact be beneficial, a natural question arises: what governs performance in this modern, over-parameterized regime? A remarkably consistent empirical finding is the existence of \emph{neural scaling laws} \cite{kaplan2020scaling, hoffmann_training_2022}, which describe a predictable, power-law relationship between a model's performance (typically measured by the cross-entropy loss, $\mathcal{L}$) and three key factors: the number of model parameters ($N$), the size of the training dataset ($P$), and the amount of compute used for training ($C \propto PN$).

The test loss is often found to scale as a power law in each of these variables when the other two are not bottlenecks. For example, for a model trained on a very large dataset, the loss is often observed to scale with the number of parameters as:
\begin{equation}
    \mathcal{L}(N) \approx \left(\frac{N_c}{N}\right)^{\alpha_N} + \mathcal{L}_\infty,
    \label{eq:scaling_law_params}
\end{equation}
where $N_c$ and $\alpha_N$ are constants specific to the model family and task, and $\mathcal{L}_\infty$ represents an irreducible error term. Similar power-law relationships hold for the dataset size $P$ if the model is large enough
\begin{equation}
    \mathcal{L}(P) \approx \left(\frac{P_c}{P}\right)^{\alpha_P} + \mathcal{L}_\infty.
    \label{eq:scaling_law_data}
\end{equation}
Furthermore, given the exponents $\alpha_N,\alpha_P$, it is often desired to balance them such that given a compute budget $C$, it is allocated such that one is on the \emph{compute optimal frontier}. The loss scaling directly implies how one should scale up
\begin{align}
    P &\approx C^\frac{\alpha_N}{\alpha_N+\alpha_P};
    &
    N &\approx C^\frac{\alpha_P}{\alpha_N+\alpha_P},
\end{align}
and indeed, LLMs today are scaled according to this prescription~\cite{hoffmann_training_2022}.

These scaling laws suggest that in the ``infinite everything'' limit, performance improvements are not exhausted but can be reliably achieved simply by increasing scale. This has profound practical implications, as it provides a predictive framework for estimating the return on investment for building larger models or collecting more data. The origin of these scaling laws is an active area of research, but a leading hypothesis is that they are not a property of the learning algorithm alone, but rather reflect the intrinsic statistical structure of the data itself \cite{geiger_scaling_2020,bahri_explaining_2021,sharma_scaling_2022,paquette20244+,bordelon_dynamical_2024,bordelon2025feature,cagnettaDerivingNeuralScaling2026}. Natural data, such as images and text, may possess a power-law structure in its own right (e.g., in its spectral density or compressibility)~\cite{levi_underlying_2025,cagnettaDerivingNeuralScaling2026}, and the scaling laws of neural networks arise from their ability to effectively learn and exploit this underlying structure. This implies that the path to better models lies not just in architectural innovation, but fundamentally in harnessing more data and computation to better capture the statistics of the world. 

Finally, we should emphasize that unlike physical laws which have centuries of precise experimental evidence to back them up, scaling ``laws'' (and their associated exponents) are not necessarily universal properties of the data. For example, the scaling exponent can depend on whether the network is in the feature learning regime or not~\cite{bordelon2025feature}, or how the data is pre-processed~\cite{Batson:2023ohn}. In fact, the functional form of the scaling laws may itself be an artifact of overly-constrained fits~\cite{barkeshli2026originneuralscalinglaws}. That said, the general phenomenology of scaling laws has important implications for choices about training data, architecture, and compute resources, regardless of the specific details.

\subsection{Solvable high-dimensional learning curves and phase transitions}
In physics-motivated settings such as teacher--student and random-feature models, one can often compute learning curves, i.e., test error versus the sample ratio $\alpha=P/N$, or related ratios) in the high-dimensional limit where $P,N$ diverge at fixed proportions, and identify sharp phase transitions, separating learnable and unlearnable regimes.
For generalized linear models and random linear estimation, statistical-physics analyses (often via replicas) can be made rigorous and yield Bayes-optimal generalization as well as precise information-theoretic thresholds; on the algorithmic side, approximate message passing admits a tractable state evolution that tracks their dynamics and pinpoints regimes of algorithmic optimality and computational-statistical gaps~\cite{barbier2019optimal,bayati_montanari_2011,rangan_gamp_2011}; recent work has made progress understanding the interpolation $\alpha \simeq 1$ regime of deep nonlinear networks with width scaling as the input dimension~\cite{barbier2025statistical}.
Parallel lines of work provide closed-form learning curves for kernel regression and random feature models, deriving learning curves through spectral alignment, effective dimension, and feature-map design~\cite{canatar_spectral_2021,Loureiro_2022,refinetti2021classifying,cohen_learning_2021,seroussi_separation_2023}.
Moving from static estimators to \emph{training dynamics}, dynamical mean-field theory (DMFT) and related high-dimensional limits yield low-dimensional ODE/SDE descriptions of (mini-batch) SGD, making explicit how learning rate, batch size, training time, and width control transitions between qualitatively different learning regimes and generalization outcomes~\cite{goldt2019dynamics,Mignacco_2021,veiga_phase_2023,arnaboldi_unifying_2023}.
Recent solvable analyses also quantify when feature learning breaks the kernel limit after finitely many steps, including explicit sample-complexity scalings for alignment with multi-index structure~\cite{dandi_giant_2024,cui_asymptotics_2024}.
Although idealized, these models are useful as they show how inductive biases affect the sample complexity required for reliable learning, and they can even reproduce power-law learning curves reminiscent of neural scaling laws in tractable limits~\cite{bordelon_dynamical_2024}.

\section{Design Choices and Their Impacts: Architecture and Hyperparameters}
\label{sec:design}
In the previous section, we discussed the benefits of scaling up the model size and dataset. In this section, we discuss \emph{how} to scale up model size, focusing in particular on how to tune hyperparameters (HPs) and architectures at scale, since hyperparameter tuning via grid search becomes increasingly expensive, and often prohibitively so, as models grow larger.

\subsection{Parametrization and Hyperparameter Transfer}

Consider the following scenario: a developer wants to find the optimal HPs $H_{\rm large}^*$ of a large model $M_{\rm large}$, but the developer does not have sufficient compute budget to run a full grid search at scale. Being optimistic, the developer decides to execute the HPs search on a small model $M_{\rm small}$ to find $H_{\rm small}^*$ and then hopes to deduce $H_{\rm large}^*$ from it. The developer tries the following recipe
\begin{equation}
    H_{\rm large}^* = H_{\rm small}^* \frac{H_{\rm large}^0}{H_{\rm small}^0}
\end{equation}
for $H^0_{(\cdot)}$ the default HPs given at initialization by PyTorch~\cite{paszke_pytorch_2019}, TensorFlow~\cite{developers_tensorflow_2025} / Keras~\cite{chollet2015keras}. Unfortunately, the parameters found by this method don't guarantee that $M_{\rm large}$ will perform better than $M_{\rm small}$. However, there are $H_{\rm large}^0,~H_{\rm small}^0$ such that $M_{\rm large}$ will outperform $M_{\rm small}$!  A notable achievement of modern theory of deep learning was the introduction of the maximal update parametrization ($\mu$P)~\cite{yang_tensor_2021,yang_tuning_2021} and its extensions~\cite{bordelon_depthwise_2023,yang_tensor_2023,ishikawa_parameterization_2023,bordelon_infinite_2024,blake_u-mup_2024,haas_boldsymbolmumathbfp2_2024,everett_scaling_2024,dey_dont_2025,qiu_scaling_2025}, these parameterizations guarantee consistent improvement when the model width $N$ is scaled up.

The $\mu$P scaling strategy and its follow-ups are essentially recipes for these $H^0$'s, with more recent works expanding the range of HPs that can be transferred, or making it applicable in more scenarios.
The core idea behind these works is ensuring that the activations, gradients and parameter updates at each layer remain $O(1)$ throughout training.
In Table~\ref{tab:parametrizations} we give a prescription for adjusting the learning rate (LR), initialization variance (Init. Var.), a layer scaling parameter (Fwd.), and weight decay (WD) when scaling up both the width and the depth of a ResNet (e.g., transformer) trained with AdamW~\cite{loshchilov_decoupled_2018}, based on~\cite{dey_dont_2025}.

\begin{table}[t]
\centering
\label{tab:sp_alpha_color}
\begin{tabular}{lcc}
\toprule
\textbf{Model HPs} & $H^*_{\rm small}$ & $H^*_{\rm large}$ \\
\midrule

First Layer. Init. Var.
& $\tunable{\sigma_{\text{base}}^2}$
& $\tunable{\sigma_{\text{base}}^2}$ \\

First Layer. LR (AdamW)
& $\tunable{\eta_{\text{base}}}$
& $\tunable{\eta_{\text{base}}}$ \\

\midrule

Hidden Init. Var.
& $\tunable{\sigma_{\text{base}}^2}$
& $\tunable{\sigma_{\text{base}}^2}\cdot \widthctl{m_N^{-1}}$ \\

Hidden LR (AdamW)
& $\tunable{\eta_{\text{base}}}$
& $\tunable{\eta_{\text{base}}}\cdot \widthctl{m_N^{-1}}\cdot$ \\

Hidden Bias LR (AdamW)
& $\tunable{\eta_{\text{base}}}$
& $\tunable{\eta_{\text{base}}}$ \\

Hidden WD (AdamW)
& $\tunable{\lambda_{\text{base}}}$
& $\tunable{\lambda_{\text{base}}}\cdot \widthctl{m_N}$ \\

\midrule

Residual Connection
& $\mathbf{Z}^l + \mathrm{Layer}^l(\mathbf{Z}^l)$
& $\mathbf{Z}^l + \depthctl{m_L^{-1}}\cdot \mathrm{Layer}^l(\mathbf{Z}^l)$ \\

\midrule

Last Layer Init. Var.
& $\tunable{\sigma_{\text{base}}^2}$
& $\tunable{\sigma_{\text{base}}^2}$ \\

Last Layer LR (AdamW)
& $\tunable{\eta_{\text{base}}}$
& $\tunable{\eta_{\text{base}}}$ \\

Last Layer Fwd.
& $\mathbf{X}^L \mathbf{W}_{\text{L}}^{\top}$
& $\mathbf{X}^L \mathbf{W}_{\text{L}}^{\top}\cdot \widthctl{m_N^{-1}}$ \\

\midrule

AdamW $\epsilon$ (Residual blocks)
& $\tunable{\epsilon_{\text{base}}}$
& $\tunable{\epsilon_{\text{base}}}\cdot \widthctl{m_N^{-1}}\cdot \depthctl{m_L^{-1}}$ \\

AdamW $\epsilon$ (First L. \& Last L.)
& $\tunable{\epsilon_{\text{base}}}$
& $\tunable{\epsilon_{\text{base}}}$ \\

\bottomrule

\end{tabular}
\caption{The first column indicates the optimal HPs found for the small model, and the second gives a prescription for the hyperparameters of the large network. 
\tunable{Tunable} hyperparameters, and \widthctl{width}/\depthctl{depth} scaling are highlighted. $\widthctl{m_N}= N_{\rm large}/N_{\rm small}$, $\depthctl{m_L}= L_{\rm large}/L_{\rm small}$ are the width and depth ratios of the models, respectively. Adapted and simplified from~\cite{dey_dont_2025}.}
\label{tab:parametrizations}
\end{table}

\subsubsection{Extensions}
A recent study showed that models trained with $\mu$P on the compute optimal frontier show remarkably similar loss dynamics across scales~\cite{qiu_scaling_2025}. This fact can be used to diagnose problems early during training and to test the implementation of $\mu$P. 
Other works have extended $\mu$P to other settings. Ref.~\cite{haas_boldsymbolmumathbfp2_2024} extended it to sharpness-aware minimization~\cite{foret_sharpness-aware_2020}, \cite{ishikawa_parameterization_2023} extended it to second-order optimizers like Shampoo~\cite{gupta_shampoo_2018} and~\cite{blake_u-mup_2024} adapted it for low-precision friendly training. Furthermore, there exists a continuous family of initializations for which NTK and $\mu$P represent special cases, which also seems to permit scaling strategies which include network depth~\cite{yaida2022metaprincipledfamilyhyperparameterscaling,dinan2023effectivetheorytransformersinitialization}. 

\subsubsection{Limitations} The most significant limitation of $\mu$P is that $M_{\rm small}$ must be trained on the same dataset as $M_{\rm large}$ will be trained on, which usually implies major over-training. A direct implication is that if $M_{\rm large}$ is on the compute optimal frontier $M_{\rm small}$ is not. 
Additionally, there is no guarantee $\mu$P will find the optimal HPs $H^*_{\rm large}$ and indeed it sometimes doesn't~\cite{everett_scaling_2024}; notwithstanding, extensive experiments show that $\mu$P finds locally optimal HPs, and that larger models perform better under $\mu$P~\cite{yang_tuning_2021,dey_cerebras-gpt_2023,everett_scaling_2024,lingle_empirical_2025,vlassis_thorough_2024}.
It should also be noted that such HP transfer requires the $M_{\rm small}$ that is not too small, such that its normalized $H^*_{\rm small}/ H^{\mu \rm P}_{\rm small}$ optimal HPs are not too far from the infinite width limit. A practical width of $\approx 500$ is often found to be sufficient~\cite{everett_scaling_2024}.

For a complementary discussion of hyperparameter transfer see~\cite{everett_scaling_2024}. Notably, they show successful hyperparameter transfer with standard parametrization when using layer-dependent learning rates. 

\subsubsection{Use in physics} The $\mu$P scaling strategy does not appear to have taken hold in physics applications, perhaps because the sizes of most models are not yet large enough to benefit from hyperparameter transfer. However, this situation may be changing as compute becomes cheaper and more accessible. Recently, Ref.~\cite{Bahl:2026jvt} studied scaling laws for MLPs and transformers trained on regression of QFT amplitudes, where $\mu$P was employed precisely because it obviated a computationally-expensive hyperparameter search. In the experimental realm, Ref.~\cite{Vigl:2026ppx} investigates the scaling of transformers for flavor tagging and derives compute-optimal scaling laws extending to the billion-parameter scale, but does not discuss the role of initialization or hyperparameter transfer.

\subsection{Depth}

The era of deep learning refers to the incredible success of \emph{deep} neural networks, with multiple hidden layers. A zero-hidden-layer neural network performing a regression or classification task is equivalent to linear or logistic regression; the intuition behind the improving performance of deeper networks is that the hidden layers of the network may learn nontrivial representations of the data, as opposed to having to pre-process these features by hand prior to performing regression. Indeed, there is some evidence that when trained on hierarchical data, the hidden layers learn progressively fine-grained features. For instance, feature visualization and representational-similarity analyses in trained vision models suggest a progression from low-level edge/texture detectors to higher-level parts and object representations across depth~\cite{bengio_representation_2013,zeiler_visualizing_2014,yosinski_understanding_2015,raghu_svcca_2017,kornblith_similarity_2019}.

That said, the role of depth is not fully understood from a theoretical perspective. For example, given a dataset, there is no firm prediction for the optimal depth to achieve a given training objective. Said another way, if one had $N$ parameters available to train, how should those parameters be allocated between depth and width? Some hints come from taking the $N \to \infty$ limit. In NTK parameterization, if we take the width of an MLP to infinity at finite depth, the network trained under gradient descent reverts to linear regression, this time on a basis of random features given by $df/d\theta_\mu$ where $f$ is the network output and $\theta_\mu$ are all of the parameters. From this point of view, the role of depth can become irrelevant for particular architecture choices; for example, all infinite-width MLP networks in NTK parameterization with ReLU activation are ``shallow''~\cite{bietti_deep_2021}; the same is not true of CNNs and transformers in general, even in the NTK limit. On the other hand, taking the depth to infinity at finite width generically leads to chaotic behavior~\cite{Roberts:2021fes}. There are some tricks for mitigating this behavior, for example initializing the weights from an ensemble of orthogonal matrices~\cite{pennington_resurrecting_2017,xiao_dynamical_2018,Day:2023fjp}, but in general a network which is deeper than it is wide does not seem to offer convincing performance benefits. 

A promising line of work seeks to understand the role of depth in both supervised and unsupervised settings through the lens of the generative model of the training data. If one assumes that real-world data is probabilistically generated from a distribution which admits a compositional and hierarchical structure (a toy example of which is the Random Hierarchy Model~\cite{cagnetta2023deep}), it can be shown analytically and empirically that learning this structure efficiently by computing correlations requires deep latent representations~\cite{patel2015probabilistic, mossel2016deep, poggio2017why, malach2018provably, schmidt2020nonparametric, cagnetta2023deep,sclocchi2024probinglatenthierarchicalstructure, favero2025compositionalgeneralizationcreativityimprove, cagnetta2024towards,cagnetta2025learningcurvestheoryhierarchically,cagnettaDerivingNeuralScaling2026,parley2026deepnetworkslearnparse}. While this approach does not yet reveal how deep networks \textit{learn hierarchical latent structure with gradient descent}, complementary works are attempting to address this problem in supervised settings~\cite{dandi2025computationaladvantagedepthlearning}. The aforementioned initialization strategies are effectively independent of any hierarchical structure present in the training data, with the optimal depth for a given small model determined empirically by hyperparameter search before scaling up. A complete theory of hyperparameters would ideally connect these scaling strategies to the data complexity.

\subsection{Optimization, Optimizers and Learning Rates}
\label{sec:optimizers}

The choice of step size (learning rate) $\eta$ is often the most consequential optimization hyperparameter in practice.
In a simple quadratic setting, stability and progress impose a curvature-dependent constraint, i.e.\ $\eta \in (0,2/\lambda_{\max}[H])$ for $\lambda_{\max}[H]$ the largest eigenvalue of the Hessian~\cite{nesterov_introductory_2004,bubeck_convex_2015}. 
For deep networks, the relevant curvature scale is both highly anisotropic and time-dependent: the spectrum of the Hessian changes over training, sharp directions can appear or disappear, and the per-layer Hessian eigenvalues can vary dramatically across layers and time.

\subsubsection{Gradient descent as a discretized gradient flow}
GD can be viewed as a discretization scheme for \emph{gradient flow}. 
Consider a loss to be minimized $\mathcal{L}(\theta)$ and with full-batch GD,
\begin{equation}
    \theta_{t+1} = \theta_t - \eta \nabla \mathcal{L}(\theta_t).
    \label{eq:gd_update}
\end{equation}
In the continuous-time limit $\eta\to 0$ with $t\eta \to \tau$, Eq.~\eqref{eq:gd_update} approaches the ODE
\begin{equation}
    \frac{d\theta(\tau)}{d\tau} = -\nabla \mathcal{L}(\theta(\tau)).
    \label{eq:grad_flow}
\end{equation}
Near a (local) minimizer $\theta^\star$, a second-order Taylor expansion gives a quadratic model
\begin{equation}
    \mathcal{L}(\theta) \approx \mathcal{L}(\theta^\star) + \tfrac12 (\theta-\theta^\star)^\top H(\theta^\star)(\theta-\theta^\star),
\end{equation}
where $H(\theta^\star)$ is the Hessian.
Writing the error $e_t=\theta_t-\theta^\star$, GD becomes the linear recursion
\begin{equation}
    e_{t+1} = (I-\eta H)e_t.
\end{equation}
If $H$ is symmetric positive definite (local minimum), convergence requires the spectral radius $\lambda_{\max}[|I-\eta H|]<1$, which is equivalent to the above-mentioned stability condition
\begin{equation}
    0<\eta<\frac{2}{\lambda_{\max}(H)}.
    \label{eq:stepsize_stability}
\end{equation}

\subsubsection{Momentum and acceleration: a quadratic calculation}
A widely used modification of GD is Polyak's heavy-ball method (``SGD with momentum'')\footnote{We note that ``momentum'' dynamics are not equivalent to the familiar momentum in physics.}. 
Introducing a velocity variable $v_t$, one writes
\begin{align}
    v_{t+1} &= \beta v_t + \nabla \mathcal{L}(\theta_t),\\
    \theta_{t+1} &= \theta_t - \eta v_{t+1},
\end{align}
with momentum coefficient $\beta\in[0,1)$.
On a one-dimensional quadratic $\mathcal{L}(\theta)=\tfrac12 \lambda \theta^2$, this yields the linear second-order recursion
\begin{equation}
    \theta_{t+1} = (1+\beta-\eta\lambda)\theta_t - \beta\,\theta_{t-1},
\end{equation}
whose characteristic polynomial $r^2-(1+\beta-\eta\lambda)r+\beta=0$ makes explicit how momentum changes the stability region and can damp oscillations along high-curvature directions.
In deep learning practice, momentum is ubiquitous and can be viewed as both an acceleration mechanism and a low-pass filter on gradient noise~\cite{sutskever_importance_2013}; recent theory also studies acceleration for non-convex optimization and saddle-point escape~\cite{jin_agd_2018}.

\subsubsection{Adaptive algorithms and preconditioning}

A basic difficulty in gradient descent is that one scalar step size must serve every direction in parameter space at once: if the loss changes very rapidly in some directions and much more slowly in others, then a step size large enough to make visible progress in the latter may be unstable in the former, while a step size chosen for stability in the former makes progress elsewhere unnecessarily small. Ideally, one would compensate for this by multiplying the gradient by the inverse Hessian, as in Newton's method, thereby using genuine second-order information to normalize the dynamics direction by direction. In large-scale problems, however, forming, storing, and inverting the Hessian is usually too expensive. One is therefore led to modify gradient descent by replacing the flow
\[
\dot{\theta}=-\nabla L(\theta)
\]
with
\[
\dot{\theta}=-P_t\nabla L(\theta),
\]
where \(P_t\) is a positive-definite operator. Such an operator is called a \emph{preconditioner}. Geometrically, this means that descent is no longer measured in the ambient Euclidean metric, but in the metric \(ds_t^2=d\theta^\top P_t^{-1}d\theta\): a well-chosen \(P_t\) reshapes locally elongated level sets into something closer to spherical, so that the dynamics can take comparably large stable steps in all directions instead of being controlled by the most restrictive one. A particularly important practical example is RMSprop \cite{TielemanHinton2012}, which uses a time-dependent diagonal choice of \(P_t\). Writing \(g_{t,i}=\partial_i L(\theta_t)\), its update is
\[
\theta_{t+1,i}=\theta_{t,i}-\eta\,\frac{g_{t,i}}{\sqrt{v_{t,i}}+\varepsilon}.
\qquad
v_{t,i}=\beta v_{t-1,i}+(1-\beta)g_{t,i}^2,
\]
Thus each coordinate is rescaled by the inverse root-mean-square size of its recent gradients. Adam \cite{kingma_adam_2014} may be viewed as RMSprop with momentum, together with a bias-correction factors, while AdamW \cite{loshchilov_decoupled_2018} decouples weight decay from the adaptive update and is a very common practical default in large-scale training. Recent work~\cite{cohenUnderstandingOptimizationDeep2024} has shown that the dynamics induced by such adaptive preconditioners approximate a second order method. Their limitation is equally geometric: because the operator \(P_t\) is diagonal, RMSprop and Adam choose a preferred basis---the coordinate basis---and only equalize step sizes along those axes, so they cannot fully remove anisotropy when the relevant curvature directions are rotated relative to the coordinates. This is one reason for the current interest in richer matrix-valued choices of \(P_t\), and more explicitly second-order or quasi-second-order methods such as Shampoo and related variants, as well as newer geometry-aware optimizers such as Muon \cite{gupta_shampoo_2018,morwaniNewPerspectiveShampoos2024,jordan_muon_2024}.

\subsubsection{Mini-batch SGD, gradient noise, and diffusion limits}
Stochastic gradient descent (SGD) replaces $\nabla \mathcal{L}(\theta_t)$ by a mini-batch estimate,
\begin{equation}
    g_t = \nabla \mathcal{L}(\theta_t) + \xi_t,\qquad \mathbb{E}[\xi_t\mid \theta_t]=0,
\end{equation}
yielding the update $\theta_{t+1}=\theta_t-\eta g_t$~\cite{robbins_stochastic_1951,bottou_optimization_2018}. 
If the mini-batch size is $B$, a common approximation is $\mathrm{Cov}(\xi_t\mid\theta_t)\approx \Sigma(\theta_t)/B$ (up to finite-population corrections), where $\Sigma(\theta)$ is the per-example gradient covariance. 
In a small-step limit, SGD can be approximated by an It\^{o} SDE of the schematic form
\begin{equation}
    d\theta(\tau) = -\nabla \mathcal{L}(\theta(\tau))\, d\tau + \sqrt{\frac{\eta}{B}}\, \Sigma(\theta(\tau))^{1/2}\, dW_\tau,
    \label{eq:sgd_sde}
\end{equation}
where $W_\tau$ is standard Brownian motion~\cite{mandt_sgd_2017}. 
In special cases (e.g.\ approximately isotropic noise), Eq.~\eqref{eq:sgd_sde} suggests an ``effective temperature'' scaling like $T\propto \eta/B$, linking learning rate and batch size to an implicit regularization mechanism that biases toward wider minima~\cite{hochreiter_flat_1997,keskar_largebatch_2016}. In those cases the equilibrium distribution is the familiar Boltzmann distribution. Relatedly, injecting specifically crafted white noise can be used to sample the Bayesian posterior~\cite{welling_bayesian_2011}. This connection has been used to argue SGD approximately samples the Bayesian posterior~\cite{mandt_sgd_2017, mingardSGDBayesianSampler2020}, and to theorize about neural networks.

\subsubsection{Batch size coupling, critical batch size, and practical scaling rules}
Because the noise scale in Eq.~\eqref{eq:sgd_sde} depends on the ratio $\eta/B$, it is often more meaningful to discuss step size jointly with batch size.
Empirically, training throughput can be improved by increasing $B$ up to a \emph{critical batch size} beyond which the number of optimization steps to reach a target loss stops decreasing significantly; this motivates hybrid strategies that increase $B$ over time or tune $(\eta,B)$ together~\cite{goyal_accurate_2017,smith_dont_2018,mccandlish_empirical_2018}.
A practical takeaway is that ``linear scaling'' rules (increase $\eta$ proportionally with $B$) can work in some regimes, but only up to a point: once gradient noise is too small, generalization can degrade unless other regularizers compensate~\cite{keskar_largebatch_2016}.

\subsubsection{Large learning rates and the edge of stability} 
Classical analysis predicts divergence when $\eta$ violates Eq.~\eqref{eq:stepsize_stability}. 
However, modern deep networks are often trained with learning rates that appear to hover near, or even transiently exceed, the local quadratic stability threshold.
This regime has been termed the \emph{edge of stability}: the top Hessian eigenvalue tends to track $2/\eta$ while training loss may be non-monotone on short time scales, yet decreases on average~\cite{cohen_edge_2021,lewkowycz_catapult_2020}. 
Recent theory suggests that operating in this regime can induce a distinct implicit regularization mechanism tied to non-smoothness and the geometry of ``minimum-loss'' manifolds~\cite{arora_edge_2022}. 
For practitioners, this supports the heuristic that ``as large as possible but stable'' learning rates can be beneficial, especially when combined with the warmup and decay schedules which are typically used in large models. For some recent work proposing optimal learning rate schedules for large models, see Refs.~\cite{liu2025neuralthermodynamiclawslarge,bordelon2026theoryoptimallearningrate,meterezAnytimePretrainingHorizonFree2026}, though the study of optimal schedules is still in early stages, and seems to suggest the optimal schedule is a data-dependent power law.

\subsubsection{Emergent late-time generalization and grokking}
A striking example of nontrivial learning dynamics is \emph{grokking}: a model first fits the training data (often reaching near-zero training loss) while test performance remains poor, and only after prolonged training does test accuracy abruptly improve~\cite{power_grokking_2022}.
This delayed generalization has been observed in algorithmic tasks and has sparked several mechanistic hypotheses, including a competition between ``memorizing'' and ``generalizing'' solutions under explicit or implicit regularization~\cite{liu_understanding_grokking_2022, noaGrokking}.
Recent solvable models show that grokking can occur even in linear estimators and in minimal classification settings, where the full long-time dynamics can be derived analytically and the grokking time can be predicted as a function of sample size, regularization, and problem geometry~\cite{levi_grokking_linear_estimators_2024,beck_grokking_edge_linear_separability_2024, bordelonGrokking}.
From the perspective of this review, the main lesson is that \emph{time} and \emph{regularization strength} can act as control parameters: operating near critical points (e.g.\ near interpolation or separability thresholds) can produce long transients and sharp dynamical transitions in generalization, which may be practically relevant in physics settings where one often trains to extremely low training error and cares about rare-event tails.

\section{Learning Under Constraints}
\label{sec:constraints}

While Sections~\ref{sec:universal}--\ref{sec:design} emphasize behaviors that emerge cleanly in large-scale regimes (and in particular along compute-optimal scaling frontiers), many physics use cases violate at least one of the assumptions that make those regimes predictive. 
A convenient way to organize ``non-ideal'' learning problems is to decompose the excess test risk into several contributions,
\begin{equation}
    \mathcal{E}_{\rm test} \;\approx\; \mathcal{E}_{\rm Bayes}
    \;+\; \mathcal{E}_{\rm approx}
    \;+\; \mathcal{E}_{\rm est}
    \;+\; \mathcal{E}_{\rm optim},
\end{equation}
corresponding (schematically) to irreducible noise, finite-capacity approximation error, finite-sample estimation error, and finite-training optimization error.
Different constraints primarily inflate different terms, and therefore suggest different mitigation strategies. 
In physics, where one often has unusually strong prior knowledge (symmetries, conservation laws, simulators, and well-defined uncertainty budgets), the dominant theme is to \emph{trade generic scale for structured inductive bias}. Taken together, these constrained regimes motivate a common principle: when one cannot ``buy'' robustness and accuracy by scaling, one should instead (a) encode as much known structure as possible, (b) amortize expensive computation when repeated inference is needed, and (c) validate carefully against physics-specific failure modes (distribution shift, rare-event sensitivity, and systematic uncertainties). A rough categorization of the various constraints one might encounter in physics data is given in Tab.~\ref{tab:constraints_summary}.

\begin{table}[t]
\centering
\begin{tabular}{p{0.2\linewidth} p{0.3\linewidth} p{0.5\linewidth}}
\toprule
\textbf{Constraint} & \textbf{Dominant error term} & \textbf{Typical mitigation strategies} \\
\midrule
Data limited & Estimation (variance) & Encode priors/structure (invariance, equivariance), transfer/self-supervised pretraining, targeted simulation/active learning, Bayesian or kernel methods. \\
\addlinespace
Parameter limited & Approximation (bias) & Structured architectures, pruning/quantization/distillation, co-design with deployment hardware, validate observable-level stability under compression. \\
\addlinespace
Compute limited & Optimization / budget & Scaling-aware compute allocation, mixed precision and efficient training, reuse pretrained models, parameter-efficient fine-tuning, amortization via surrogates/emulators. \\
\addlinespace
Time limited & Latency / non-stationarity & Low-latency deployment (FPGAs/ASICs), small and quantized models, rapid fine-tuning workflows, monitoring for drift and controlled retraining. \\
\bottomrule
\end{tabular}
\caption{\label{tab:constraints_summary} A schematic map from common physics constraints to the dominant statistical failure mode and typical mitigation strategies. The mapping is approximate; in practice multiple constraints often apply simultaneously.}
\end{table}

\subsection{Data Limited}
Physics is often ``data limited'' in a sense that differs from Internet-scale machine learning: labels can be intrinsically sparse in the training data (e.g.\ rare processes), expensive (high-fidelity simulation), or entangled with systematic effects that reduce the \emph{effective} sample size. 
In such regimes, one may not be able to rely on scaling laws alone; instead, a reliable lever is to reduce estimation error by capitalizing on prior knowledge: symmetries, conservation laws, locality, and known detector structure, either via augmentation or via architectures with built-in invariances/equivariances~\cite{cohen_group_2016, bronstein_geometric_2021}. 
Kernel and Gaussian-process viewpoints make this tradeoff explicit: they express a relatively simple prior over function space, which can be understood in terms of eigenfunctions and eigenvalues of the model's kernel. In these models, the alignment between the model and target determines performance~\cite{canatar_spectral_2021}; the optimal choice is, of course, a kernel with a single non-zero eigenvalue and eigenfunction that is the target. In general, a limited amount of data will restrict the number of learnable modes, and the unlearnable modes act as noise, increasing the variance~\cite{cohen_learning_2021,atanasovScalingRenormalizationHighdimensional2025}.

In practice, ``data limited'' can mean several distinct (and sometimes simultaneous) situations in physics:
\begin{itemize}
    \item \textbf{Rare classes / rare processes:} the training sample for a signal process is intrinsically small (e.g.\ a rare decay channel or a low-rate background), and the relevant decision boundary must be learned from few labeled examples. This was the context of the 2020 LHC Olympics anomaly-detection challenge~\cite{Kasieczka:2021xcg}.
    \item \textbf{Expensive labels:} labels requiring high-fidelity simulation (e.g.\ detailed detector simulation, expensive lattice calculations, or high-precision numerical solvers) \
    \item \textbf{Distribution shift between simulation and data:} simulation may produce large training sets, but systematic mismatch limits how much of that data is effectively usable for generalization to real measurements. A recent example from high-energy physics is Refs.~\cite{Benato:2024lnj,Chakkappai:2025noy}.
\end{itemize}

Fundamentally, learning and generalizing from small amounts of data requires additional knowledge on the problem. The most explicit way of encoding this knowledge is by using a Bayesian prior, or choosing a problem-informed similarity measure - a kernel.
Concretely, when data are scarce and uncertainty quantification is central, Gaussian processes and kernel methods can be attractive because they make the prior explicit and often behave predictably in the small-$n$ regime~\cite{rasmussen_gaussian_2006}. 
Moreover, the kernel viewpoint provides a bridge to the infinite-width limits discussed earlier: in regimes where a neural network behaves approximately like kernel regression, the generalization behavior can be analyzed in terms of the kernel spectrum and its alignment with the task\cite{cohen_learning_2021,canatar_spectral_2021,simon_eigenlearning_2023,lavie_demystifying_2025,karkadaPredictingKernelRegression2025} (cf.\ Sec. \ref{subsec:linear_regression}).

Relatedly, symmetries, the workhorse of physics, can be used to improve learning in data scarce regime as we highlight below. Other approaches include leveraging unlabeled data and using active learning when labeled data is expansive.

\subsubsection{Encoding symmetries and structure}
One of the most powerful ways to reduce estimation error is to restrict the hypothesis class to functions that respect known symmetries.
This can be done explicitly via data augmentation, or more efficiently via \emph{equivariant} architectures that build the symmetry in by construction (so that every layer transforms predictably under the symmetry action)~\cite{cohen_group_2016,bronstein_geometric_2021}.
Many physics problems naturally come with permutation symmetries (sets of particles), rotational symmetries (detector or celestial coordinates), or Lorentz symmetries (relativistic kinematics), e.g. \cite{bogatskiyLorentzGroupEquivariant2020,spinnerLorentzEquivariantGeometricAlgebra2024,spinnerLorentzLocalCanonicalization2025}. 
As a simple example, the Deep Sets construction~\cite{zaheer_deep_2017} provides a universal form for permutation-invariant functions of sets, and has inspired a large family of set- and graph-based architectures for jet tagging and event classification. 
In jet physics, for instance, architectures that respect permutation invariance and locality can achieve strong performance with relatively modest parameter counts compared to generic MLPs; concrete examples include Energy Flow Networks~\cite{komiske_energy_2019} and ParticleNet~\cite{qu_particle_2020}. A complimentary perspective is provided in ~\cite{brehmer_does_2025} where they show empirically that equivariance may offer a trade-off between slightly worse scaling exponent and a large improvement is the constant coefficients coefficient of the loss.

\subsubsection{Leveraging unlabeled data: pretraining, self-supervision, and weak supervision}
Physics experiments often have abundant \emph{unlabeled} data even when labeled data are limited; for example, correlations among the hadrons which make up jets in high-energy physics experiements encode properties of QCD even without conditioning on the identity of the progenitor parton. A practical route is to pretrain representations using self-supervised objectives on large unlabeled corpora (possibly including simulation), and then fine-tune on the small labeled set. 
This can reduce variance by constraining the downstream model to reuse a representation learned from a broader distribution or to transfer across different datasets with similar structures (see Refs.~\cite{Mikuni:2025ocp,Elsharkawy:2026kwp} for some recent efforts along these lines using point-cloud data). Relatedly, weak supervision (noisy labels, proxy tasks, or physics-based heuristics) can be used to trade label precision for label volume, provided that the resulting systematic uncertainties are propagated.

\subsubsection{Active learning and targeted simulation}
When labels are expensive because they require simulation or expert time, \emph{active learning} can be used to adaptively select which points to label next, focusing resources on the regions of feature space that most reduce uncertainty~\cite{settles_active_2010}. 
In physics, this idea often appears as \emph{targeted simulation}: generate additional Monte Carlo only in kinematic corners that dominate the analysis sensitivity or where the current model is least certain. 
The central statistical principle is to spend limited labeling/simulation budget where it maximally reduces estimation error.

\subsubsection{Example: group averaging (augmentation) as variance reduction}
To illustrate how symmetry can translate into sample efficiency, suppose the data distribution is invariant under a finite group $G$ acting on inputs (e.g.\ rotations), and labels satisfy $y(x)=y(g\!\cdot\! x)$.
Given a predictor $f$, define its \emph{group-averaged} (invariant) version
\begin{equation}
    \bar f(x)\;=\;\frac{1}{|G|}\sum_{g\in G} f(g\!\cdot\! x).
\end{equation}
For convex losses $\ell(\cdot,y)$, Jensen's inequality implies
\begin{equation}
    \ell(\bar f(x),y)\;\le\;\frac{1}{|G|}\sum_{g\in G}\ell(f(g\!\cdot\! x),y),
\end{equation}
so training on an augmented dataset (or using an equivariant architecture whose output is invariant) can be viewed as a principled way to \emph{average out} nuisance directions, reducing variance without worsening the best achievable augmented empirical loss.
While this does not automatically guarantee better test performance, it makes explicit why symmetry constraints can be especially valuable in small-$n$ regimes~\cite{cohen_group_2016,bronstein_geometric_2021}.

\subsection{Parameter Limited}

A ``parameter-limited'' regime arises when the model must be small because of memory constraints, power constraints, or a hard limit on model size for deployment (e.g.\ on-detector inference, FPGA/ASIC triggers, or embedded systems). 
Statistically, this constraint primarily increases \emph{approximation error}: even with infinite data and perfect optimization, the hypothesis class may be too restricted to represent the target function at the required fidelity. Practically speaking, this type of error can be mitigated by training models in the overparamterized regime, which is known both theoretically and empirically to be beneficial for convergence of the dynamics to generalizing solutions, and subsequently performing distillation to a substantially smaller model.

\subsubsection{Inductive bias \& parameter efficiency}
A key lesson from physics is that incorporating the right structure can reduce parameter requirements by orders of magnitude.
Convolutions, message passing, and equivariant layers are all examples of architectures that trade generic capacity for structured function classes, often improving both sample efficiency and parameter efficiency.

\subsubsection{Compression: pruning, quantization, and distillation}
When one starts from an accurate but over-parameterized model, a common strategy is to compress it while maintaining performance:
\begin{itemize}
    \item \textbf{Pruning} removes weights or channels that contribute little to the output, ideally yielding sparse or structured models that run faster on hardware~\cite{han_learning_2015}. This approach is useful for the inference stage, for fitting a large trained model on a smaller end device.
    \item \textbf{Quantization} reduces numerical precision (e.g.\ INT8/INT4) to lower memory bandwidth and accelerate inference; quantization-aware training can substantially reduce the accuracy drop compared to post-training quantization~\cite{jacob_quantization_2018}. The quantization step can be done either before and throughout training, or post-hoc for an already trained model.
    \item \textbf{Knowledge distillation} trains a smaller ``student'' network to match the softened outputs of a larger ``teacher,'' often transferring decision boundaries more effectively than training the student from scratch~\cite{hinton_distilling_2015}. This approach requires either an already trained model for the student model or training both the student and the teacher from scratch. 
\end{itemize}
These techniques can be combined (e.g.\ distill first, then quantize), but the ordering matters because each step can change the effective function class and therefore the optimal hyperparameters.

\subsubsection{Example: low-latency inference in collider triggers}
A canonical physics example is real-time event selection at colliders, where inference must run under tight latency and resource budgets.
Frameworks like \texttt{hls4ml} translate trained neural networks into FPGA firmware, enabling sub-microsecond inference while maintaining a transparent mapping between the ML model and the deployed hardware representation~\cite{duarte_fast_2018}. 
In this setting, parameter count, arithmetic precision, and inference-time latency are coupled constraints; successful deployment often requires training with quantization and resource constraints ``in the loop,'' followed by dedicated validation to ensure that compression does not introduce analysis-relevant biases.

\subsubsection{Compression and Occam's razor and generalization}
Beyond deployment constraints, compression can also serve as a \emph{post hoc} indicator of effective model complexity.
A representative result is that if a trained network can be compressed (e.g.\ by pruning/quantization) to $b$ bits while preserving training performance, then the generalization gap can be bounded on the order of $\tilde O(\sqrt{b/n})$ (hiding log factors), making explicit why heavily overparameterized models can still generalize well when their learned weights are ``compressible''~\cite{arora_compression_2018}. 
This connects parameter-limited deployment directly to generalization: small compressed size, implies an upper bound on the minimal description length (MDL) and on the Kolmogorov complexity is often thought of as a formal Occam's razor. Related to our discussion above, it has been claimed that larger models results in more compressible solutions~\cite{goldblumNoFreeLunch2024a}, explaining why distillation is sometimes preferred to training a small model on the raw data.

\subsubsection{The lottery tickets hypothesis}
A related empirical phenomenon is the \emph{lottery ticket hypothesis}: within a large randomly initialized network there often exist sparse subnetworks that, when trained in isolation (often with a suitable initialization), match the accuracy of the full model~\cite{frankle_lottery_2019}. 
This suggests an alternative workflow: train a large model to discover a good representation, identify an efficient subnetwork (via pruning or structured sparsification), and then retrain or fine-tune the subnetwork under the deployment constraints. Along this line,~\cite{buzagloHowUniformRandom2025} shows that if there exists a narrow network that can express the target function, a typical larger model will generalize well.

\subsection{Compute Limited}
A ``compute-limited'' regime refers to constraints on the total training budget (FLOPs, accelerator hours, or energy), which restricts either (i) the model size that can be trained, (ii) the amount of data that can be processed, or (iii) the number of optimization steps that can be taken. At a practical level, compute is always the eventual limiting factor: roughly speaking, compute scales with both data and parameters, so for a finite dataset and a finite number of parameters, any frontier model is compute-limited by wherever training stops.

\subsubsection{Compute allocation and scaling-aware design}
When scaling laws are available, they provide a principled way to allocate compute between parameters, data, and training time (cf.\ Sec.~\ref{sec:scalinglaws}). 
The compute-optimal perspective exemplified by Chinchilla-style scaling arguments~\cite{hoffmann_training_2022} is directly useful in physics settings where one can estimate the marginal return of additional compute and decide whether to spend it on more simulation, a larger model, or longer training.

\subsubsection{Algorithmic and systems-level efficiency}
Many compute savings come from engineering rather than new estimators: mixed-precision training reduces memory bandwidth and increases throughput with minimal loss in accuracy~\cite{micikevicius2017mixed}; gradient checkpointing trades compute for memory; and distributed training can improve wall-clock time if communication overhead is controlled.
From a statistical viewpoint, these techniques matter because they expand the reachable region of $(N,P,C)$ without changing the learning objective.

\subsubsection{Transfer learning and parameter-efficient adaptation}
A common way to escape compute limits is to avoid training from scratch. 
Pretraining on large, generic corpora (possibly including simulation) and then fine-tuning on a smaller physics dataset can move a problem closer to the ``large-data'' regime at dramatically reduced cost.
For very large pretrained models, \emph{parameter-efficient fine-tuning} methods such as low-rank adaptation (LoRA) can update only a small number of additional parameters while keeping the base model frozen~\cite{hu_lora_2021}. 
This reduces both compute and memory footprint, and is particularly attractive when physics groups wish to adapt a shared foundation model to multiple downstream tasks.

\subsubsection{Amortization and surrogate modeling}
Compute limits also motivate amortizing expensive computations across many downstream queries.
Examples include emulators of expensive simulations, fast surrogates for likelihood functions, and amortized inference networks that replace repeated iterative fits with a single trained model.
In these cases, one pays a one-time training cost to reduce the marginal cost of subsequent analyses; the statistical challenge shifts toward ensuring that the surrogate preserves the relevant uncertainty structure and does not introduce uncontrolled bias.

\subsubsection{Early stopping as compute-aware (and statistically meaningful) regularization}
When compute limits prevent full convergence, the \emph{training trajectory} itself becomes part of the estimator.
In least-squares and kernel settings, gradient descent with early stopping is known to act as an implicit regularizer: stopping after $T$ iterations corresponds to a spectral filtering of the data and can achieve near-optimal bias--variance tradeoffs without explicit penalty terms~\cite{yao_earlystop_2007}. 
Practically, this means that under compute constraints, tuning \emph{how long} to train (and how to schedule $\eta_t$) is not only an optimization decision but also a statistical one: it chooses an effective regularization strength that can reduce overfitting when data are limited and can mitigate distribution shift by preventing overly sharp fits to simulation artifacts.

\subsection{Time Limited}

``Time limited'' can refer to at least two distinct constraints that are common in physics:
(i) \textbf{strict inference latency}, where a decision must be made within microseconds--milliseconds (online triggers, real-time monitoring, low-latency astrophysical alerts), and
(ii) \textbf{rapid turnaround}, where models must be trained, validated, and updated on short cycles as detector conditions, calibrations, or analysis selections evolve.

\subsubsection{Low-latency inference}
Low-latency settings couple statistical and hardware concerns: the effective hypothesis class is constrained not only by parameter count, but also by the allowable arithmetic, memory access patterns, and pipeline depth.
As discussed above, FPGA deployment frameworks such as \texttt{hls4ml} make these constraints explicit and therefore enable end-to-end co-design of models and hardware~\cite{duarte_fast_2018}. 
From a generalization standpoint, the key risk is that aggressive compression or low precision can shift decision boundaries in ways that correlate with physics observables; this motivates validation protocols that explicitly monitor analysis-relevant distributions under deployment constraints.

\subsubsection{Rapid updates and non-stationarity}
Many experiments operate in a non-stationary environment: detector conditions drift, calibrations change, and the data distribution can evolve.
This calls for workflows that support continual evaluation and periodic retraining or fine-tuning, while guarding against catastrophic forgetting and uncontrolled feedback loops.
The broader machine-learning literature on concept drift and streaming data provides useful frameworks and diagnostics~\cite{gama_survey_2014}, but physics adds the requirement that updates must remain interpretable within an uncertainty budget and consistent with established calibration procedures.

\section{Conclusion}

In this article, we have surveyed many of the key features of deep learning, with the goal of justifying at least some of the bewilderingly large set of seemingly-arbitrary choices which confront any practitioner when starting to build and deploy a deep learning model. While much remains to be understood from first-principles theory, we hope that we have illustrated the benefit of a physics style of reasoning in understanding phenomena such as benign overfitting, neural scaling laws, optimizer and hyperparameter choices, and the tradeoffs between prior knowledge and raw data volume. Unlike many applications of deep learning, physics data demands a level of rigor which may motivate a certain set of choices to achieve a strong inductive bias, possibly at the expense of ``performance'' as measured simply by a lower loss over the course of training. The field of ``physics for AI'' in the deep learning era is still in its infancy, and we are confident that further research along the lines we have sketched in this article will bring tangible benefits to the ``AI for physics'' community. 

\section*{Acknowledgments}

We thank the VERaiPHY team, especially Gaia Grosso and Ramon Winterhalder, for helpful discussions, as well as Sascha Diefenbacher and Maria Ubaldi for comments which improved the presentation of this article. The authors thank the Aspen Center for Physics, which is supported by National Science Foundation grant
PHY-2210452, where portions of this work were completed. Y.K. acknowledges the support of a Discovery Grant from the
Natural Sciences and Engineering Research Council of Canada (NSERC) and the support of the Canadian AI Safety Institute Research Program at CIFAR. Resources used in
preparing this research were provided, in part, by the Province of Ontario, the Government of Canada through CIFAR,
and companies sponsoring the Vector Institute.

\clearpage

\bibliography{references}

\end{document}